\documentclass{article}

\usepackage{microtype}
\usepackage{graphicx}
\usepackage{subfigure}
\usepackage{booktabs} 

\usepackage{hyperref}

\usepackage[accepted]{icml2018}

\usepackage{times}
\usepackage{epsfig}
\usepackage{graphicx}
\usepackage{amsmath}
\usepackage{amssymb}

\usepackage{color}
\usepackage[normalem]{ulem}
\usepackage{calligra}
\usepackage{multirow}

\definecolor{mayablue}{rgb}{0.45, 0.76, 0.98}
\definecolor{burntorange}{rgb}{0.8, 0.33, 0.0}

\newcommand{\RR}{\mathbb{R}}

\newcommand{\picwThree}{0.32\linewidth}

\newcommand{\mySkip}[1]{{\hskip #1\linewidth}}

\newcommand{\eg}{\textit{e.g.}}
\newcommand{\ie}{\textit{i.e.}}
\newcommand{\etc}{\textit{etc.}}

\icmltitlerunning{End-to-end Active Object Tracking via Reinforcement Learning}

\begin{document}

\twocolumn[
\icmltitle{End-to-end Active Object Tracking via Reinforcement Learning}



\icmlsetsymbol{equal}{*}

\begin{icmlauthorlist}
\icmlauthor{Wenhan Luo}{equal,tai}
\icmlauthor{Peng Sun}{equal,tai}
\icmlauthor{Fangwei Zhong}{pku}
\icmlauthor{Wei Liu}{tai}
\icmlauthor{Tong Zhang}{tai}
\icmlauthor{Yizhou Wang}{pku}
\end{icmlauthorlist}

\icmlaffiliation{tai}{Tencent AI Lab}
\icmlaffiliation{pku}{Peking University}

\icmlcorrespondingauthor{Wenhan Luo}{whluo.china@gmail.com}
\icmlcorrespondingauthor{Peng Sun}{pengsun000@gmail.com}
\icmlcorrespondingauthor{Fangwei Zhong}{zfw@pku.edu.cn}
\icmlcorrespondingauthor{Wei Liu}{wl2223@columbia.edu}
\icmlcorrespondingauthor{Tong Zhang}{tongzhang@tongzhang-ml.org}
\icmlcorrespondingauthor{Yizhou Wang}{yizhou.Wang@pku.edu.cn}

\icmlkeywords{Machine Learning, ICML}

\vskip 0.3in
]



\printAffiliationsAndNotice{\icmlEqualContribution} 

\begin{abstract}
We study active object tracking, where a tracker takes as input the visual observation (\ie, frame sequence) and produces the camera control signal (\eg, move forward, turn left, \etc). 
Conventional methods tackle the tracking and the camera control separately, which is challenging to tune jointly. It also incurs many human efforts for labeling and many expensive trial-and-errors in real-world. 
To address these issues, we propose, in this paper, an end-to-end solution via deep reinforcement learning, where a ConvNet-LSTM function approximator is adopted for the direct frame-to-action prediction. We further propose an environment augmentation technique and a customized reward function, which are crucial for a successful training. 
The tracker trained in simulators (ViZDoom, Unreal Engine) shows good generalization in the case of unseen object moving path, unseen object appearance, unseen background, and distracting object. It can restore tracking when occasionally losing the target. With the experiments over the VOT dataset, we also find that the tracking ability, obtained solely from simulators, can potentially transfer to real-world scenarios. 

\end{abstract}

\section{Introduction}
\label{sec:intro}
\begin{figure}[t]
\label{fig:pipeline}
\begin{center}
\includegraphics[width=0.85\linewidth]{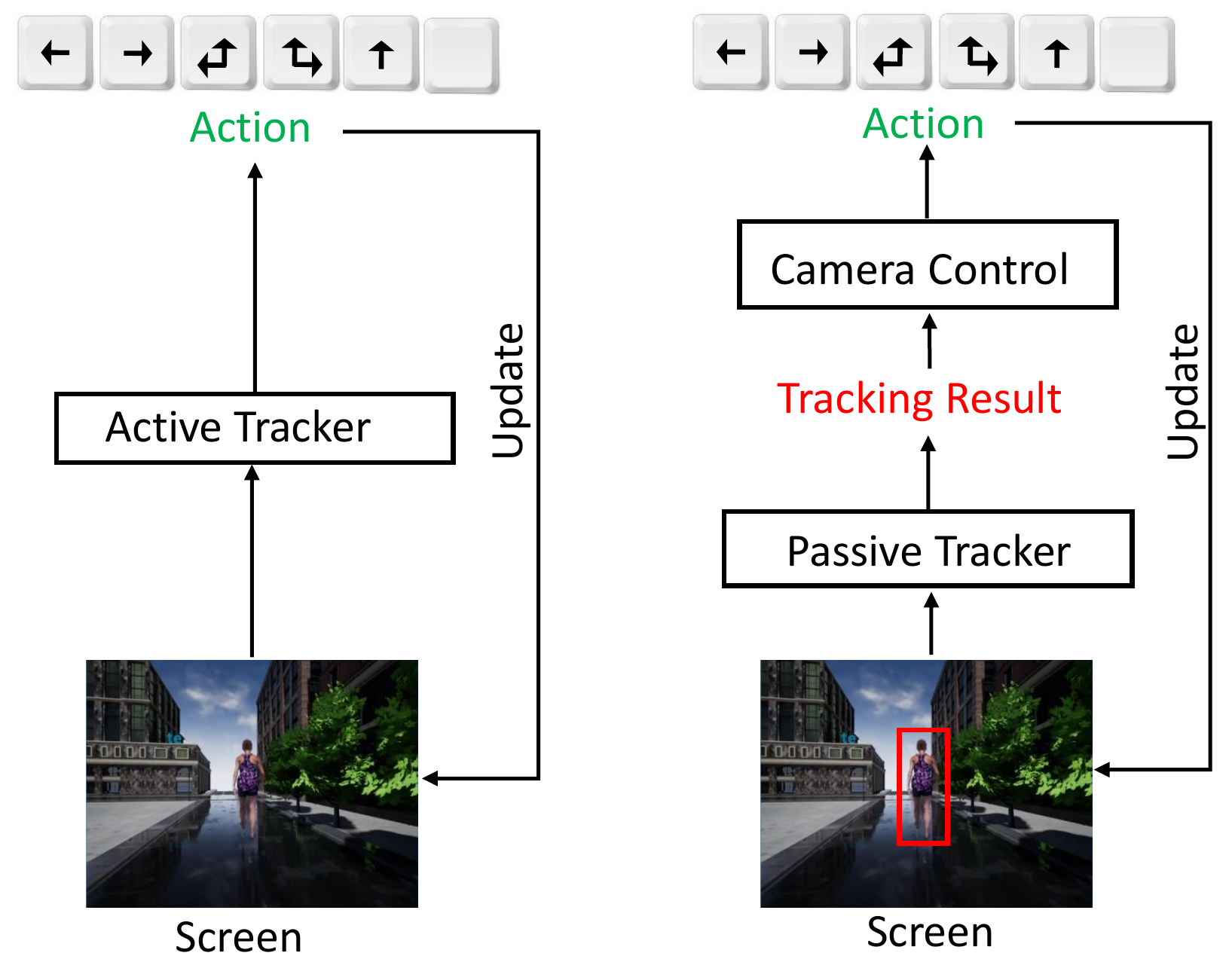}
\end{center}
\caption{The pipeline of active tracking. Left: end-to-end approach. Right: passive tracking plus other modules.}
\end{figure}

Object tracking has gained much attention in recent decades \cite{bertinetto2016staple,Danelljan_2017_CVPR,zhu2016beyond,cui2016recurrently}. The aim of object tracking is to localize an object in continuous video frames given an initial annotation in the first frame. 
Much of the existing work is, however, on the \emph{passive} tracker, where it is presumed that the object of interest is always in the image scene so that there is no need to handle the camera control during tracking. This fashion is inapplicable to some use-cases, \eg, the tracking performed by a mobile robot with a camera mounted or by a drone.
To this end, one should seek a solution of \emph{active} tracking, which composes two sub-tasks, \ie, the object tracking and the camera control (Fig. \ref{fig:pipeline}, Right).

Unfortunately, it is hard to jointly tune the pipeline with the two separate sub-tasks. The tracking task may also involve many human efforts for bounding box labeling. Moreover, the implementation of camera control is non-trivial and can incur many expensive trial-and-errors happening in real-world. To address these issues, we propose an end-to-end active tracking solution via deep reinforcement learning. To be specific, we adopt a ConvNet-LSTM network, taking as input raw video frames and outputting camera movement actions (\eg, move forward, turn left, \etc). 

We leverage virtual environments to conveniently simulate active tracking, saving the expensive human labeling or real-world trial-and-error. In a virtual environment, an agent (\ie, the tracker) observes a state (a visual frame) from a first-person perspective and takes an action, and then the environment returns the updated state (next visual frame). We adopt the modern Reinforcement Learning (RL) algorithm A3C \cite{mnih2016asynchronous} to train the agent, where a customized reward function is designed to encourage the agent to be closely following the object.

We also adopt an environment augmentation technique to boost the tracker's generalization ability. For this purpose, much engineering is devoted to preparing various environments in different object appearances, different backgrounds, and different object trajectories. We manage this by either using a simulator's plug-in or developing specific APIs to communicate with a simulator engine. See Sec. \ref{sec:tracking_scenario}.

To our slight surprise, the trained tracker shows good generalization capability. In testing, it performs robust active tracking in the case of unseen object movement path, unseen object appearance, unseen background, and distracting object. 
Additionally, the tracker can restore tracking when it occasionally loses the target due to, \eg, abrupt object movement.


In our experiments, the proposed tracking approach also outperforms a few representative conventional passive trackers which are equipped with a hand-tuned camera-control module. While we are not pursuing a state-of-the-art passive tracker in this work, the experimental results do show that a passive tracker is not indispensable in active tracking. Alternatively, a direct end-to-end solution can be effective. As far as we know, there has not yet been any attempt to deal with active tracking in an end-to-end manner.

Finally, we perform qualitative evaluation on some video clips taken from the VOT dataset \cite{VOT_TPAMI}. The results show that the tracking ability, obtained purely from simulators, can potentially transfer to real-world scenarios.


\section{Related Work}
\label{sec:relatedwork}

\textbf{Object Tracking.} Roughly, object tracking \cite{wu2013online} is conducted in both passive and active ways. As mentioned in Sec. \ref{sec:intro}, passive object tracking has gained more attention due to its relatively simpler problem settings. In recent decades, passive object tracking has achieved a great progress \cite{wu2013online}. Many approaches \cite{hu2012single} have been proposed to overcome difficulties resulted from the issues such as occlusion and illumination variations. In \cite{ross2008incremental} subspace learning was adopted to update the appearance model of an object and integrated into a particle filter framework for object tracking. Babenko \textit{et al.} \cite{babenko2009visual} employed multiple instance learning to track an object. Correlation filter based object tracking \cite{Valmadre_2017_CVPR,Choi_2017_CVPR} has also achieved a success in real-time object tracking \cite{bolme2010visual,henriques2015high}. In \cite{hare2016struck}, structured output prediction was used to constrain object tracking, avoiding converting positions to labels of training samples. 
In \cite{kalal2012tracking}, Tracking, Learning and Detection (TLD) were integrated into one framework for long-term tracking, where a detection module can re-initialize the tracker once a missing object reappears. 
Recent years have witnessed the success of deep learning in object tracking \cite{wang2016stct,bertinetto2016fully}. For instance, a stacked autoencoder was trained to learn good representations for object tracking in \cite{wang2013learning}. Both low-level and high-level representations were adopted to gain both accuracy and robustness \cite{ma2015hierarchical}.       
 
Active object tracking additionally considers camera control compared with traditional object tracking. There exists not much research attention in the area of active tracking. Conventional solutions dealt with object tracking and camera control in separate components \cite{denzler1994active,murray1994motion,kim2005detecting}, but these solutions are difficult to tune. Our proposal is completely different from them as it tackles object tracking and camera control simultaneously in an end-to-end manner.

\begin{figure*}[t]
\begin{center}
\includegraphics[width=0.85\linewidth]{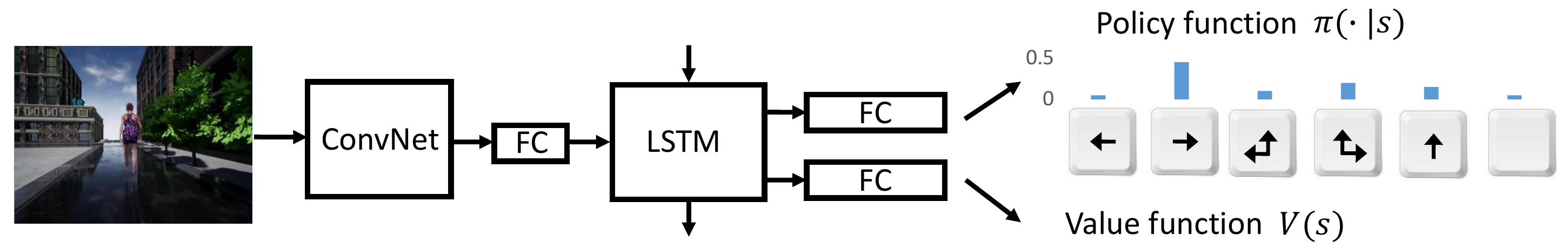}
\end{center}
\caption{The architecture of the ConvNet-LSTM network.}
\label{fig:network}
\end{figure*}

\textbf{Reinforcement Learning.} Reinforcement Learning (RL) \cite{sutton1998} intends for a principled approach to temporal decision making problems. In a typical RL framework, an \emph{agent} learns from the \emph{environment} a \emph{policy} function that maps \emph{state} to \emph{action} at each discrete time step, where the objective is to maximize the accumulated \emph{rewards} returned by the environment. 
Historically, RL has been successfully applied to inventory management, path planning, game playing, \etc

On the other hand, the past half decade has witnessed a breakthrough in deep learning applied to computer vision tasks, including image classification \cite{alexnet}, segmentation \cite{long2015}, object detection and localization \cite{rcnn}, and so on. In particular, researchers believe that deep Convolutional Neural Networks (ConvNets) can learn good features from raw image pixels, which is able to benefit higher-level tasks. 

Equipped with deep ConvNets, RL also shows impressive successes on those tasks involving image (-like) raw states, \eg, playing board game GO \cite{alphago2016} and video game \cite{atari2015, wu2017}. Recently, in the computer vision community there are also preliminary attempts of applying deep RL to traditional tasks, \eg, object localization \cite{caicedo2015} and region proposal \cite{jie2016}. There are also methods of visual tracking relying on RL \cite{choi2017visual,Huang_2017_ICCV,Supancic_2017_ICCV,Yun_2017_CVPR}. However, they are distinct from our work, as they formulate passive tracking with RL but have nothing to do with camera control. While our focus in this work is active tracking.   

\section{Our Approach}
\label{sec:approach}
In our approach, virtual tracking scenes are generated for both training and testing. To train the tracker, we employ a state-of-the-art reinforcement learning algorithm, A3C \cite{mnih2016asynchronous}. 
For the sake of robust and effective training, we also propose data augmentation techniques and a customized reward function, which are elaborated later.

Although various types of states are available, for a research purpose we let the state be only an RGB screen frame of the first-person perspective in this study. To be more specific, the tracker observes the raw visual state and takes one action from the action set $\mathcal{A} = \lbrace$\emph{turn-left, turn-right, turn-left-and-move-forward, turn-right-and-move-forward, move-forward, no-op}$\rbrace$. The action is processed by the environment, which returns to the agent the updated screen frame as well as the current reward.

\subsection{Tracking Scenarios}
\label{sec:tracking_scenario}
It is impossible to train the desired end-to-end active tracker in real-world scenarios. Thus, we adopt two types of virtual environments for simulated training. 

\textbf{ViZDoom.} ViZDoom \cite{kempka2016vizdoom,ViZDoom} is an RL research platform based on a 3D FPS video game called Doom. In ViZDoom, the game engine corresponds to the environment, while the video game player corresponds to the agent. The agent receives from the environment a state and a reward at each time step. 
In this study, we make customized ViZDoom maps (see Fig. \ref{fig:map-screen}) composed of an object (a monster) and background (ceiling, floor, and wall). The monster walks along a pre-specified path programmed by the ACS script \cite{kempka2016vizdoom}, and our goal is to train the agent, \ie, the tracker, to follow closely the object.  

\textbf{Unreal Engine.} Though convenient for research, ViZDoom does not provide realistic scenarios. To this end, we adopt Unreal Engine (UE) \cite{unrealengine} to construct nearly real-world environments. UE is a popular game engine and has a broad influence in the game industry. It provides realistic scenarios which can mimic real-world scenes (please see exemplar images in Fig. \ref{fig:unreal-env} and videos in our supplementary materials). We employ UnrealCV \cite{qiu2017unrealcv}, which provides convenient APIs, along with a wrapper \cite{gymunrealcv2017} compatible with OpenAI Gym \cite{opanai_gym}, for interactions between RL algorithms and the environments constructed based on UE. 

\subsection{A3C Algorithm}
\label{sec:actor-critic}
Following \cite{mnih2016asynchronous}, we adopt a popular RL algorithm called Actor-Critic. At time step $t$, we denote by $s_t$ the observed state, which corresponds to a raw RGB frame. The action set is denoted by $\mathcal{A}$ of size $K = |\mathcal{A}|$. An action, $a_t \in \mathcal{A}$, is drawn from a policy function distribution: $a_t \backsim \pi(\cdot|s_t) \in \RR^K$, referred to as an \emph{Actor}. The environment then returns a reward $r_t \in \RR$ according to a \emph{reward function} $r_t = g(s_t)$, which will be characterized in Sec. \ref{sec:reward-fun}. The updated state $s_{t+1}$ at next time step $t+1$ is subject to a certain but unknown state transition function $s_{t+1} = f(s_t, a_t)$, governed by the environment. In this way, we can observe a \emph{trace} consisting of a sequence of tuplets $\tau = \lbrace \ldots, \left(s_t, a_t, r_t\right) , \left( s_{t+1}, a_{t+1}, r_{t+1}\right), \ldots \rbrace$. Meanwhile, we denote by $V(s_t) \in \RR$ the expected accumulated reward in the future given state $s_t$ (referred to as \emph{Critic}).


The policy function $\pi\left( \cdot\right) $ and the value function $V\left( \cdot\right)$ are then jointly modeled by a neural network, as will be discussed in Sec. \ref{sec:network}. Rewriting them as $\pi(\cdot|s_t;\theta)$ and $V(s_t; \theta')$ with parameters $\theta$ and $\theta'$, respectively, we can learn $\theta$ and $\theta'$ over the trace $\tau$ with simultaneous stochastic policy gradient and value function regression:
\begin{align}
\label{eq:param-update}
	& \theta \leftarrow \theta + \alpha \big(R_t - V\left(s_t\right)\big) \nabla_{\theta} \log\pi\left(a_t|s_t\right) + \beta \nabla_{\theta} H\big(\pi\left(\cdot|s_t\right)\big),  \\
	& \theta' \leftarrow \theta' - \alpha \nabla_{\theta'} \frac{1}{2} \big(R_t - V\left(s_t\right)\big)^2,
\end{align}
where $R_t = \sum_{t'=t}^{t+T-1} \gamma^{t'-t} r_{t'}$ is a discounted sum of future rewards up to $T$ time steps with factor $0 < \gamma \le 1$, $\alpha$ is the learning rate, $H\left( \cdot\right)$ is an entropy regularizer, and $\beta$ is the regularizer factor.

During training, several threads are launched, each maintaining an independent environment-agent interaction. However, the network parameters are shared across the threads and updated every $T$ time steps asynchronously in a lock-free manner using Eq. (\ref{eq:param-update}) in each thread. This kind of many-thread training is reported to be fast yet stable, leading to improved generalization \cite{mnih2016asynchronous}. Later in Sec. \ref{sec:aug-env}, we will introduce environment augmentation techniques to further improve the generalization ability.
\begin{figure}
\begin{center}
\includegraphics[width=0.7\linewidth]{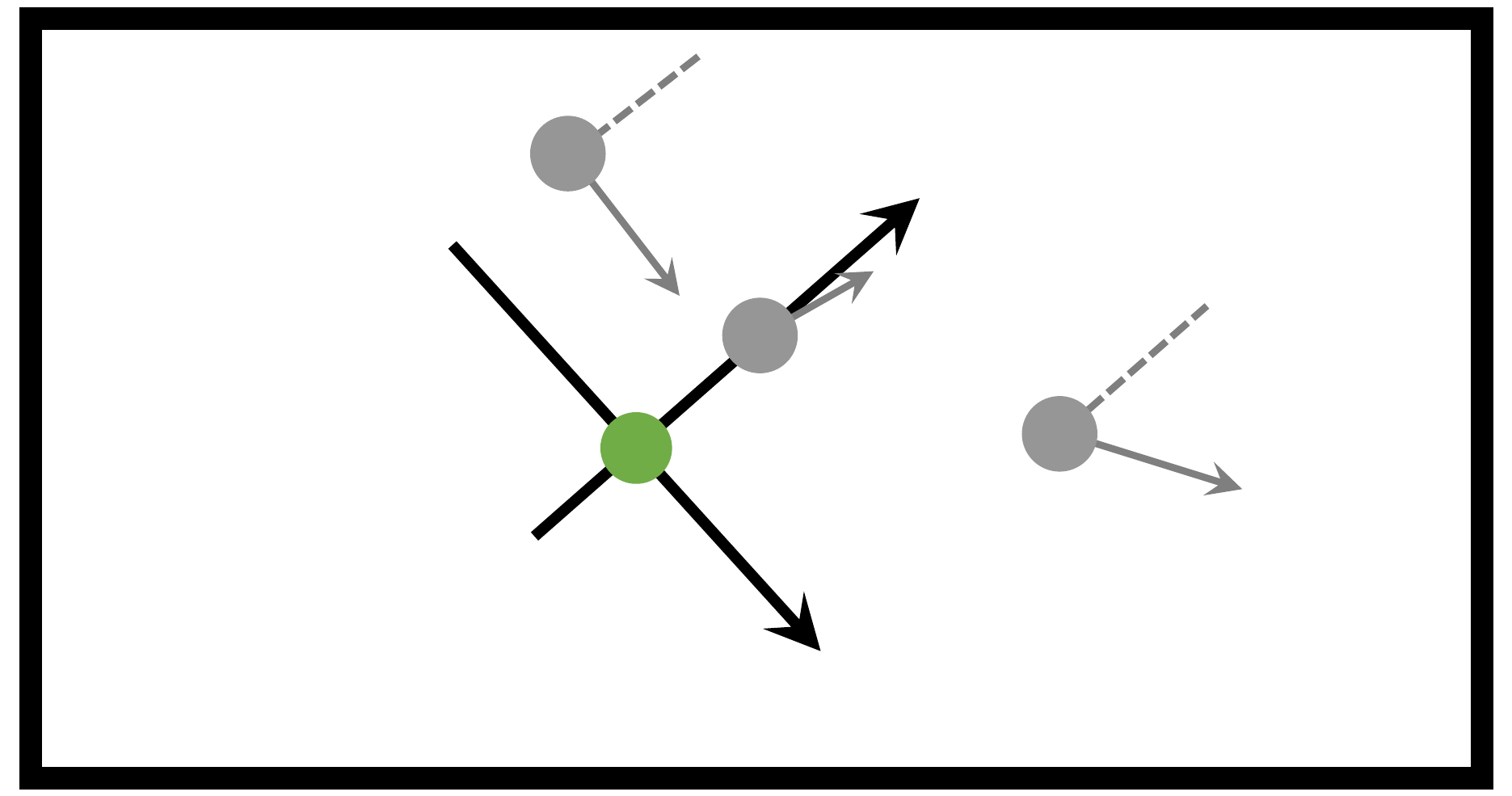}
\end{center}
\vspace{-0.4cm}
\caption{A top view of a map with the local coordinate system. The green dot indicates the agent (tracker). The gray dot indicates the initial position and orientation of an object to be tracked. Three gray dots mean three possible initial configurations. Arrow indicates the orientation of an object. Dashed gray lines are parallel to the $y$-axis. The outer thick black rectangle represents the boundary. Best viewed in color.}
\label{fig:coord}
\end{figure}

\subsection{Network Architecture}
\label{sec:network}
The tracker is a ConvNet-LSTM neural network as shown in Fig.  \ref{fig:network},
where the architecture specification is given in the following table.
The FC6 and FC1 correspond to the 6-action policy $\pi\left( \cdot|s_t\right)$ and the value $V(s_t)$, respectively. The screen is resized to $84 \times 84 \times 3$  RGB image as the network input. \\
\begin{scriptsize}
\raggedright
\begin{tabular}{c|c|c|c|c|c}
\hline
Layer\# & 1 & 2 & 3 & 4 & 5 \\
\hline
\multirow{2}{*}{Parameters} & \multirow{2}{*}{C8$\times$8-16\emph{S}4} & \multirow{2}{*}{C4$\times$4-32\emph{S}2} & \multirow{2}{*}{FC256} & \multirow{2}{*}{LSTM256} & FC6  \\
\cline{6-6}
 & & & & & FC1 \\
 \hline
\end{tabular}
\end{scriptsize}

   

\subsection{Reward Function}
\label{sec:reward-fun}
To perform active tracking, it is a natural intuition that the reward function should encourage the agent to closely follow the object. In this line of thought, firstly we define a two-dimensional local coordinate system, denoted by $\mathcal{S}$ (see Fig. \ref{fig:coord}). The $x$-axis points from the agent's left shoulder to right shoulder, and the $y$-axis is perpendicular to the $x$-axis and points to the agent's front. The origin is where the agent is. System $\mathcal{S}$ is parallel to the floor. Secondly, we manage to obtain object's local coordinate $(x, y)$ and orientation $a$ (in radius) with regard to system $\mathcal{S}$.

With a slight abuse of notation, we can now write the reward function as 
\begin{equation}
\label{eq:reward}
	r = A - \left( \frac{\sqrt{x^2 + (y - d)^2}}{c} + \lambda |a| \right),
\end{equation}
where $A > 0$, $c > 0$, $d > 0$, $\lambda > 0$ are tuning parameters. In plain English, Eq. (\ref{eq:reward}) says that the maximum reward $A$ is achieved when the object stands perfectly in front of the agent with a distance $d$ and exhibits no rotation (see Fig. \ref{fig:coord}).

In Eq. (\ref{eq:reward}) we have omitted the time step subscript $t$ without loss of clarity. Also note that the reward function defined in this way does not explicitly depend on the raw visual state. Instead, it depends on certain internal states. Thanks to the APIs provided by virtual environments, we are able to access the interested internal states and develop the desired reward function.

\vspace{-0.2cm}
\begin{figure}[t]
\hspace*{0.01\linewidth} RandomizedSmall   \mySkip{0.075} CacoDemon  \mySkip{0.065} SharpTurn \\
\includegraphics[width=\picwThree]{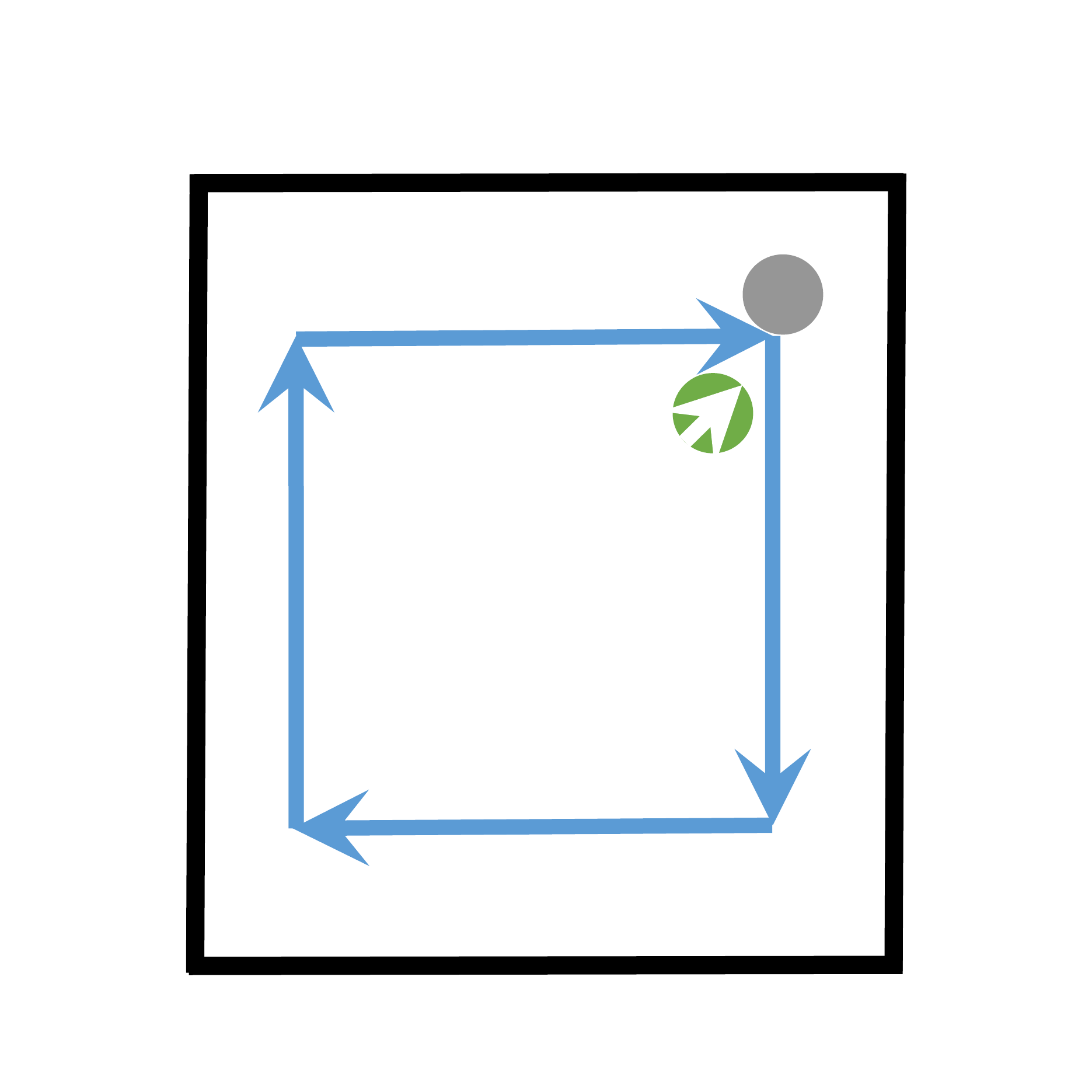}
\includegraphics[width=\picwThree]{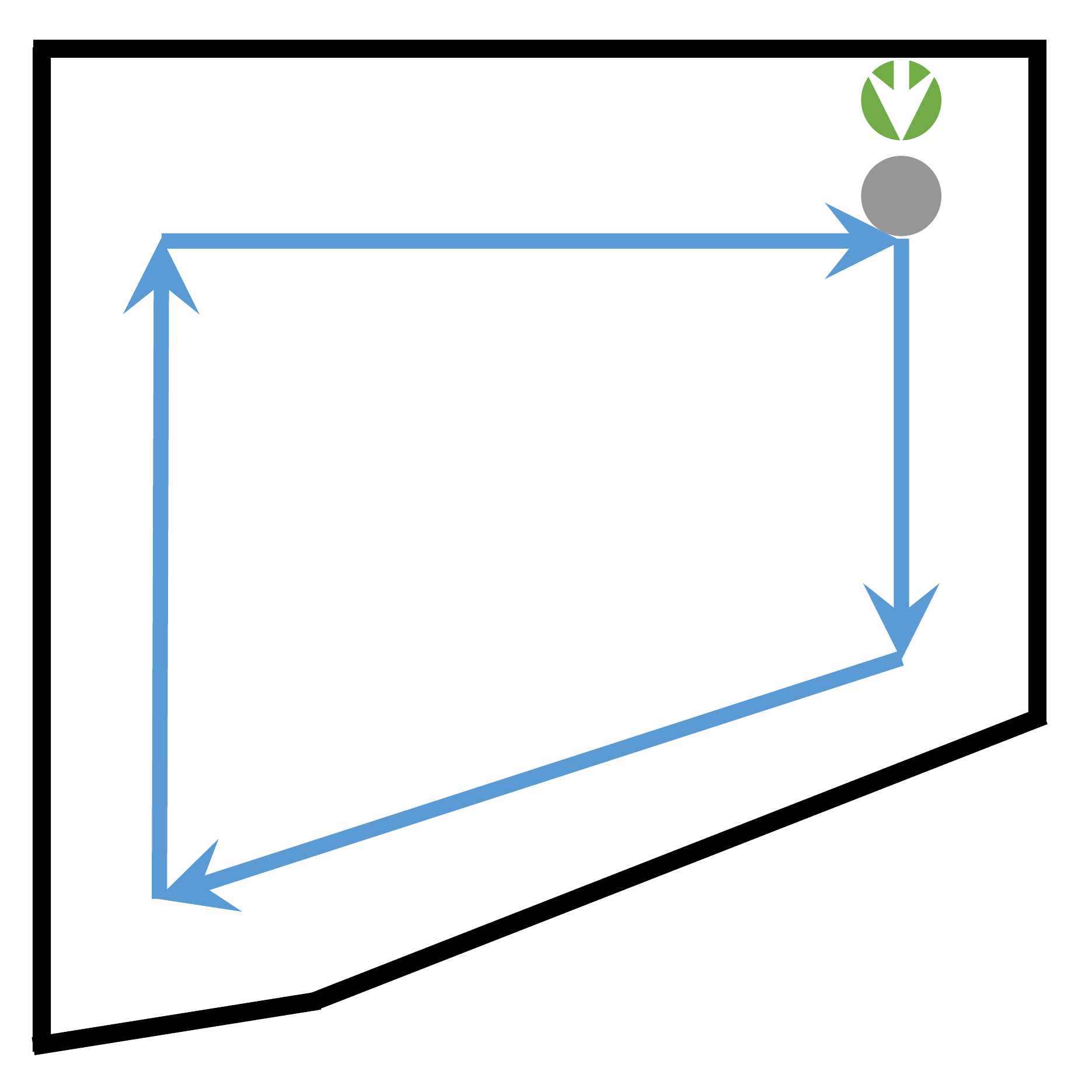} 
\includegraphics[width=\picwThree]{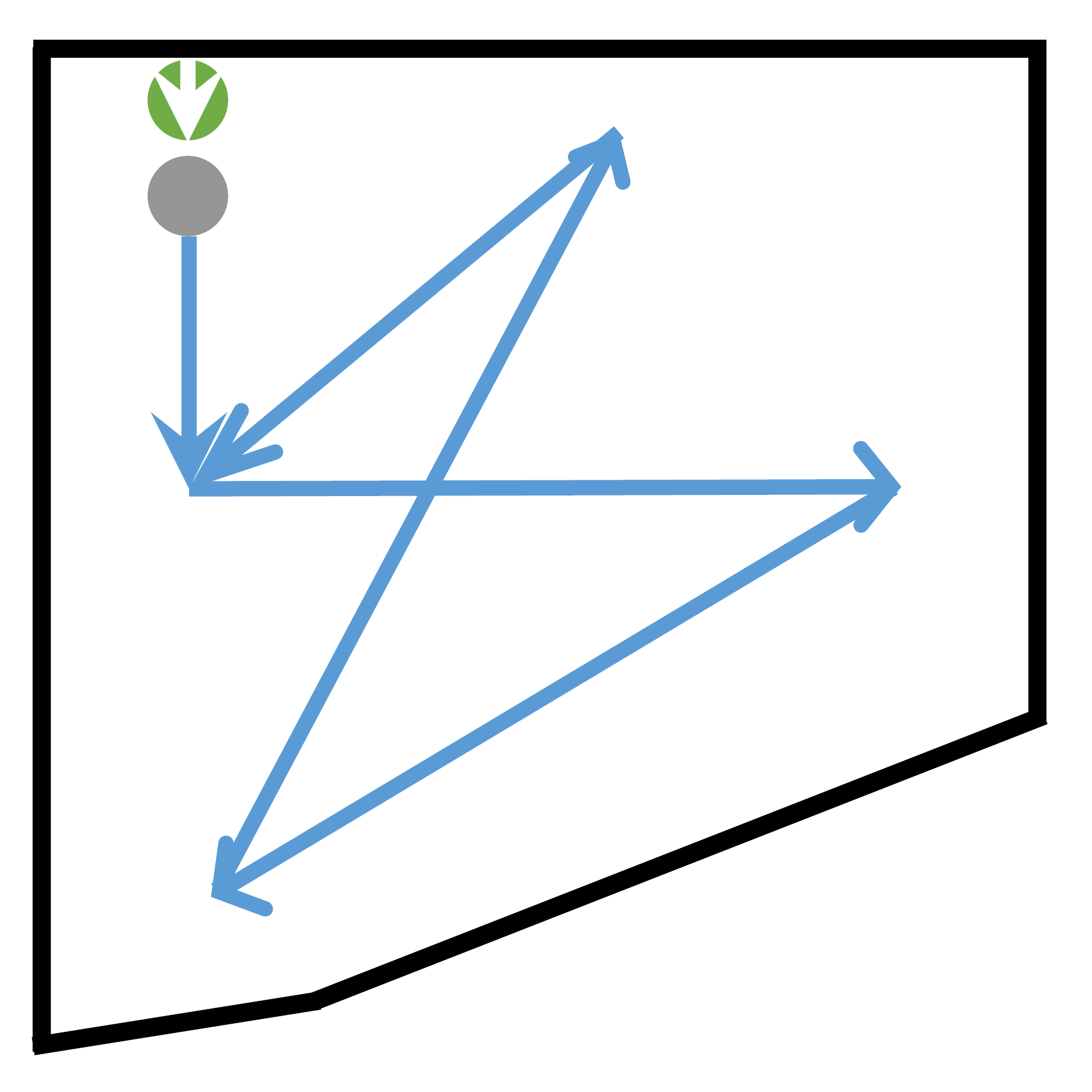}\\
\includegraphics[width=\picwThree]{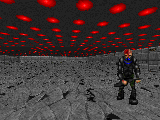}
\includegraphics[width=\picwThree]{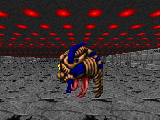}
\includegraphics[width=\picwThree]{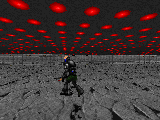}\\
\vspace{-0.4cm}
\caption{Maps and screenshots of ViZDoom environments. In all maps, the green dot (with white arrow indicating orientation) represents the agent. The gray dot indicates the object. Blue lines are planned paths and black lines are walls. Best viewed in color.}
\label{fig:map-screen}
\vspace{-0.4cm}
\end{figure}

\begin{figure*}[t]
\begin{center}
\includegraphics[width=0.95\linewidth]{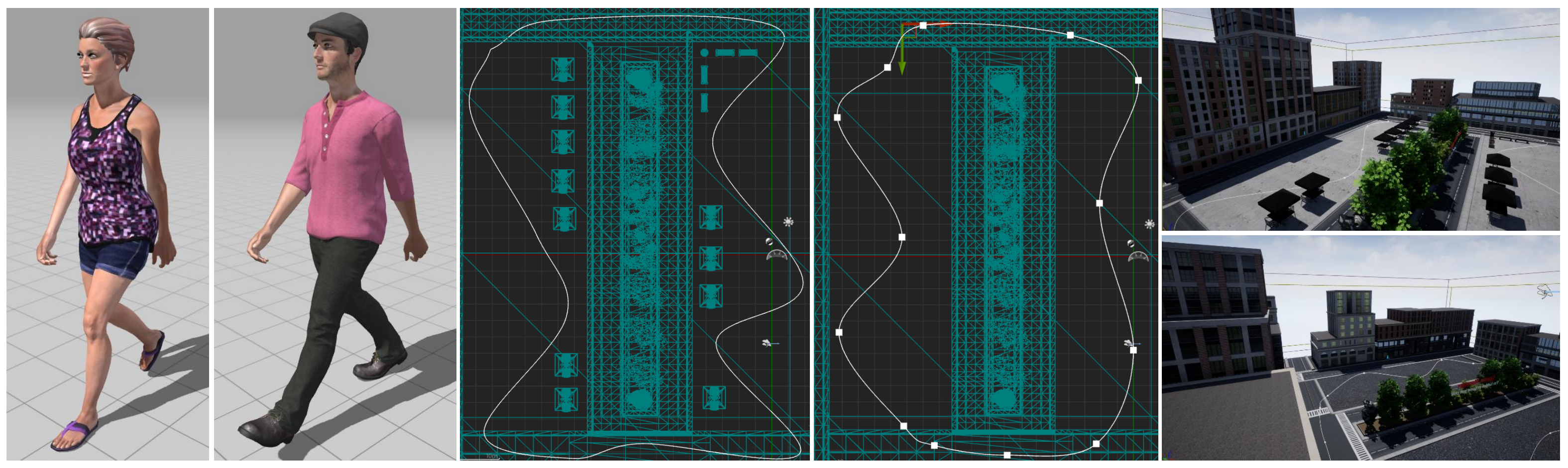}\\
\end{center}
\vspace{-0.4cm}
\caption{Screenshots of UE environments. From left to right, there are \emph{Stefani}, \emph{Malcom}, \emph{Path1}, \emph{Path2}, \emph{Square1} and \emph{Square2}. Best viewed in color.}
\label{fig:unreal-env}
\vspace{-0.4cm}
\end{figure*}

\subsection{Environment Augmentation}
\label{sec:aug-env}
To make the tracker generalize well, we propose simple yet effective techniques for environment augmentation during training. 

For ViZDoom, recall the object's local position and orientation $(x, y, a)$ in system $\mathcal{S}$ described in Sec. \ref{sec:reward-fun}. For a given environment (\ie, a ViZDoom map) with initial $(x, y, a)$,  we randomly perturb it $N$ times by editing the map with the ACS script \cite{kempka2016vizdoom}, yielding a set of environments with varied initial positions and orientations $\{x_i, y_i, a_i\}_{i=1}^N$. We further allow flipping left-right the screen frame (and accordingly the left-right action). As a result, we obtain $2N$ environments out of one environment. See Fig. \ref{fig:coord} for an illustration of several possible initial positions and orientations in the local system $\mathcal{S}$. During the A3C training, we uniformly randomly sample one of the $2N$ environments at the beginning of every episode. As will be seen in Sec. \ref{subsec:vizdoom_exp}, this technique significantly improves the generalization ability of the tracker.


 
For UE, we construct an environment with a character/target walking following a fixed path. To augment the environment, we randomly choose some background objects (\eg, tree or building) in the environment and make them invisible. At the same time, every episode starts from the position, where the agent fails at the last episode. This makes the environment and the starting point different from episode to episode, so the variations of the environment during training are augmented.

\section{Experimental Results}
\label{sec:experiment}
The settings are described in Sec. \ref{sec:settings}. The experimental results are reported for the virtual environments ViZDoom (Sec. \ref{subsec:vizdoom_exp}) and UE (Sec. \ref{subsec:ue_exp}). Qualitative evaluation is performed for real-world scenarios taken from the VOT dataset (Sec. \ref{subsec:real-world-exp}). To investigate what the tracker has learned, we conduct ablation analysis using a saliency visualization technique \cite{simonyan2013deep} in Sec. \ref{subsec:saliency}.

\subsection{Settings}
\label{sec:settings}


\textbf{Environment.}
A set of environments are produced for both training and testing.
For ViZDoom, we adopt a training map as in Fig. \ref{fig:map-screen}, left column. This map is then augmented as described in Sec. \ref{sec:aug-env} with $N=21$, leading to $42$ environments that we can sample from during training. For testing, we make other 9 maps, some of which are shown in Fig. \ref{fig:map-screen}, middle and right columns. In all maps, the path of the target is pre-specified, indicated by the blue lines. However, it is worth noting that the object does not strictly follow the planned path. Instead, it sometimes randomly moves in a ``zig-zag'' way during the course, which is a built-in game engine behavior. This poses an additional difficulty to the tracking problem.


For UE, we generate an environment named \emph{Square} with random invisible background objects and a target named \emph{Stefani} walking along a fixed path for training. For testing, we make another four environments named as \emph{Square1StefaniPath1 (S1SP1)}, \emph{Square1MalcomPath1 (S1MP1)}, \emph{Square1StefaniPath2 (S1SP2)}, and \emph{Square2MalcomPath2 (S2MP2)}. As shown in Fig. \ref{fig:unreal-env}, \emph{Square1} and \emph{Square2} are two different maps, \emph{Stefani} and \emph{Malcom} are two characters/targets, and \emph{Path1} and \emph{Path2} are different paths. Note that, the training environment \emph{Square} is generated by hiding some background objects in \emph{Square1}.

For both ViZDoom and UE, we terminate an episdoe when either the accumulated reward drops below a threshold or the episode length reaches a maximum number. In our experiments, we let the reward threshold be -450 and the maximum length be 3000, respectively.

\textbf{Metric.}
Two metrics are employed for the experiments. Specifically, \emph{Accumulated Reward} (AR) and \emph{Episode Length} (EL) of each episode are calculated for quantitative evaluation. 
Note that, the immediate reward defined in Eq. (\ref{eq:reward}) measures the goodness of tracking at some time step, so the metric AR is conceptually much like \emph{Precision} in the conventional tracking literature. 
Also note that too small AR means a failure of tracking and leads to a termination of the current episode. As such, the metric EL roughly measures the duration of good tracking, which shares the same spirit of the metric \emph{Successfully Tracked Frames} in conventional tracking applications. 
When letting $A = 1.0$ in Eq. (\ref{eq:reward}), we have that the theoretically maximum AR and EL are both 3000 due to our episode termination criterion. 
In all the following experiments, 100 episodes are run to report the mean and standard deviation, unless specified otherwise. 

\textbf{Implementation details.}
\label{subsec:implementation}
We include the implementation details in the supplementary material due to the space constraint.

\subsection{Active Tracking in The ViZDoom Environment}
\label{subsec:vizdoom_exp}
We firstly test the active tracker in a testing environment named \emph{Standard}, showing the effectiveness of the proposed environment augmentation technique. The second part is contributed to the experiments in more challenging testing environments which vary from the \emph{Standard} environment with regard to object appearance, background, path, and object distraction. Comparison with a set of traditional trackers is conducted in the last part.

\textbf{Standard Testing Environment.}
In Tab. \ref{table:diff_protocol}, we report the results in an independent testing environment named \emph{Standard} (see supplementary materials for its detailed description), where we compare two training protocols: with (called \emph{RandomizedEnv}) or without (called \emph{SingleEnv}) the augmentation technique as in Sec. \ref{sec:aug-env}. As can be seen, \emph{RandomizedEnv} performs significantly better than \emph{SingleEnv}. 


\begin{figure}[t]
\begin{center}
\includegraphics[width=0.99\linewidth]{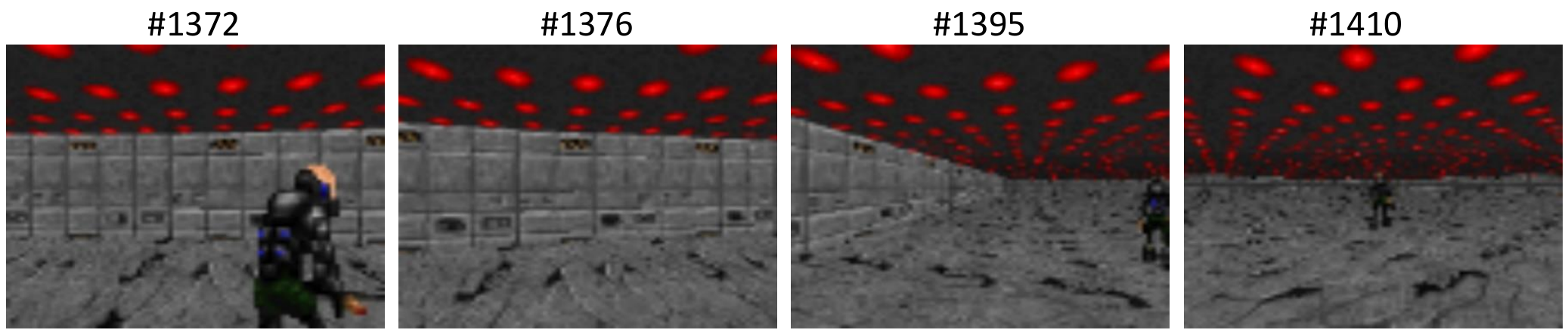}
\end{center}
\vspace{-0.4cm}
\caption{Recovering tracking when the target disappears in the \emph{SharpTurn} environment.}
\label{fig:sharpturn}
\vspace{-0.5cm}
\end{figure}

\begin{table}
\caption{Performance of different protocols in the \emph{Standard} testing environment.}
\label{table:diff_protocol}
\centering
{\begin{tabular}{c c c c}
\hline
\textbf{Protocol} & \textbf{AR} & \textbf{EL}\\
\hline
\emph{RandomizedEnv}       &2547$\pm$58  &2959$\pm$32  \\
\emph{SingleEnv}   &840$\pm$602  &2404$\pm$287 \\
\hline
\end{tabular}
}
\vspace{-0.5cm}
\end{table}


We discover that the \emph{SingleEnv} protocol quickly exhausts the training capability and obtains the best validation result at about $9 \times 10^6$ training iterations. On the contrary, the best validation result of \emph{RandomizedEnv} protocol occurs at $48 \times 10^6$, showing that the capacity of the network is exploited better despite the longer training time. In the following experiments, we only report experimental results with the \emph{RandomizedEnv} protocol.

\textbf{Various Testing Environments.} 
To evaluate the generalization ability of our active tracker, we test it in 8 more challenging environments as in Tab. \ref{table:diff_cases}. Comparing to the training environment, they present different target appearances, different backgrounds, more varying paths, and distracting targets. See supplementary materials for the detailed description.

\begin{table}[t]
\caption{Performance of the proposed active tracker in different testing environments.}
\label{table:diff_cases}
\centering
{\begin{tabular}{c c  c}
\hline
\textbf{Environment} & \textbf{AR} & \textbf{EL}\\
\hline
\emph{CacoDemon} & 2415$\pm$71  & 2981$\pm$10 \\
\emph{Zombie}    & 2386$\pm$86  & 2904$\pm$40 \\
\hline
\emph{FloorCeiling} & 1504 $\pm$ 158  & 2581 $\pm$ 84 \\
\emph{Corridor}     & 2636 $\pm$ 34   & 2983 $\pm$ 17 \\
\hline
\emph{SharpTurn}        & 2560$\pm$34  &2987$\pm$12 \\
\emph{Counterclockwise} & 2537$\pm$58  &2964$\pm$23 \\
\hline
\emph{Noise1} & 2493$\pm$72   &2977$\pm$14 \\
\emph{Noise2} & 2313$\pm$103  &2855$\pm$56 \\
\hline
\end{tabular}
}
\vspace{-0.5cm}
\end{table}

From the 4 categories in Tab. \ref{table:diff_cases} we have findings below.\\
1) The tracker generalizes well in the case of target appearance changing (\emph{Zombie}, \emph{Cacodemon}).\\
2) The tracker is insensitive to background variations such as changing the ceiling and floor (\emph{FloorCeiling}) or placing additional walls in the map (\emph{Corridor}).\\
3) The tracker does not lose a target even when the target takes several sharp turns (\emph{SharpTurn}). Note that in conventional tracking, the target is commonly assumed to move smoothly.\\ 
We also observe that the tracker can recover tracking when it accidentally loses the target. As shown in Fig. \ref{fig:sharpturn}, the target turns right suddenly and the tracker loses it (frame \#1372). Although the target completely disappears in the image, the tracker takes a series of \emph{turn-right} actions (frame \#1376 to \#1394). It rediscovers the target (frame \#1410), and continues to track steadily afterwards. 
We believe that this capability attributes to the LSTM unit which takes into account historical states when producing current outputs.\\
Our tracker performs well when the target walks counterclockwise (\emph{Counterclockwise}),  indicating that the tracker does not work by simply memorizing the turning pattern.\\
4) The tracker is insensitive to a distracting object (\emph{Noise1}), even when the ``bait'' is very close to the path (\emph{Noise2}). 

The proposed tracker shows satisfactory generalization in various unseen environments. Readers are encouraged to watch more result videos provided in our supplementary materials.

\begin{table}[t]
\caption{Comparison with traditional trackers. The best results are shown in bold.}
\label{table:comparison_thirdparty}
\centering
{\begin{tabular}{c c c c}
\hline
\textbf{Environment} & \textbf{Tracker} & \textbf{AR} & \textbf{EL}\\
\hline
\multirow{5}{*}{\emph{Standard}}
& MIL         & -454.2 $\pm$ 0.3  & 743.1 $\pm$ 21.4 \\
& Meanshift   & -452.5 $\pm$ 0.2  & 553.4 $\pm$ 2.2  \\
& KCF         & -454.1 $\pm$ 0.2  & 228.4 $\pm$ 5.5  \\
& Correlation & -453.6 $\pm$ 0.2 & 252.7 $\pm$ 16.6 \\
\cline{2-4}
& Active      & \textbf{2457$\pm$58}  & \textbf{2959$\pm$32}  \\
\hline
\multirow{5}{*}{\emph{SharpTurn}}& MIL & -453.3 $\pm$ 0.2   & 388.3 $\pm$ 15.5 \\
& Meanshift & -454.4 $\pm$ 0.3 & 250.1 $\pm$ 1.9\\
& KCF &  -452.4 $\pm$ 0.2  & 199.2 $\pm$ 5.7 \\
&Correlation & -453.0 $\pm$ 0.2 & 186.3 $\pm$ 6.0\\
\cline{2-4}
&Active      & \textbf{2560$\pm$34}     & \textbf{2987$\pm$12}  \\
\hline
\multirow{5}{*}{\emph{Cacodemon}}& MIL & -453.5 $\pm$ 0.2   & 540.6 $\pm$ 18.2 \\
& Meanshift & -452.9 $\pm$ 0.2  & 484.3 $\pm$ 9.4\\
& KCF &  -454.5 $\pm$ 0.3  &  263.1 $\pm$ 6.2 \\
&Correlation & -453.3 $\pm$ 0.2 & 155.8 $\pm$ 1.9\\
\cline{2-4}
&Active & \textbf{2451$\pm$71}    & \textbf{2981$\pm$10}  \\
\hline
\end{tabular}
}
\vspace{-0.5cm}
\end{table}

\textbf{Comparison with Simulated Active Trackers.} In a more extensive experiment we compare the proposed tracker with a few traditional trackers. These trackers are originally developed for passive tracking applications. Particularly, the MIL \cite{babenko2009visual}, Meanshift \cite{comaniciu2000real}, KCF \cite{henriques2015high}, and Correlation \cite{danelljan2014accurate} trackers are employed for comparison. We implement them by directly invoking the interface from OpenCV \cite{OpenCV} (MIL, KCF and Meanshift trackers) and Dlib \cite{Dlib} (Correlation tracker).

To make the comparison feasible, we add to the passive tracker an additional PID-like module for the camera control, enabling it to interact with the environment (see Fig. \ref{fig:pipeline}, Right). 
In the first frame, a manual bounding box must be given to indicate the object to be tracked. For each subsequent frame, the passive tracker then predicts a bounding box, which is passed to the ``Camera Control'' module. Finally, the action is produced by ``pulling back'' the target to its position in a previous frame (see supplementary materials for the details of the implementation).
For a fair comparison with the proposed active tracker, we employ the same action set $\mathcal{A}$ as described in Sec. \ref{sec:approach}.

Armed with this camera-control module, the performance of traditional trackers is compared with the active tracker in \emph{Standard}, \emph{SharpTurn} and \emph{Cacodemon}. The results in Tab. \ref{table:comparison_thirdparty} show that the end-to-end active tracker beats the simulated ``active'' trackers by a significant gap. We investigate the tracking process of these trackers and find that they lose the target soon. The Meanshift tracker works well when there is no camera shift between continuous frames, while in the active tracking scenario it loses the target soon. Both KCF and Correlation trackers seem not capable of handling such a large camera shift, so they do not work as well as the case in passive tracking. The MIL tracker works reasonably in the active case, while it easily drifts when the object turns suddenly. 

Recalling Fig. \ref{fig:sharpturn}, another reason of our tracker beating the traditional trackers is that our tracker can quickly discover the target again in the case that it is missed. While the simulated active trackers can hardly recover from failure cases.



%

\vspace{-0.2cm}
\begin{figure}[t]

\includegraphics[width=0.99\linewidth]{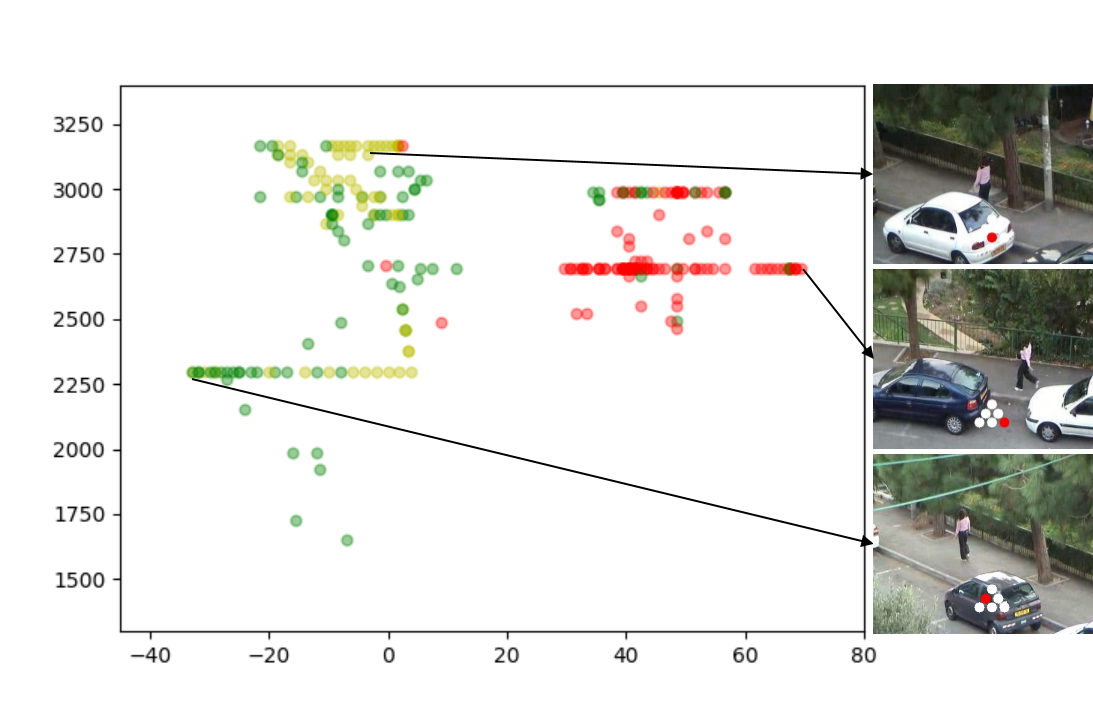}
\includegraphics[width=0.99\linewidth]{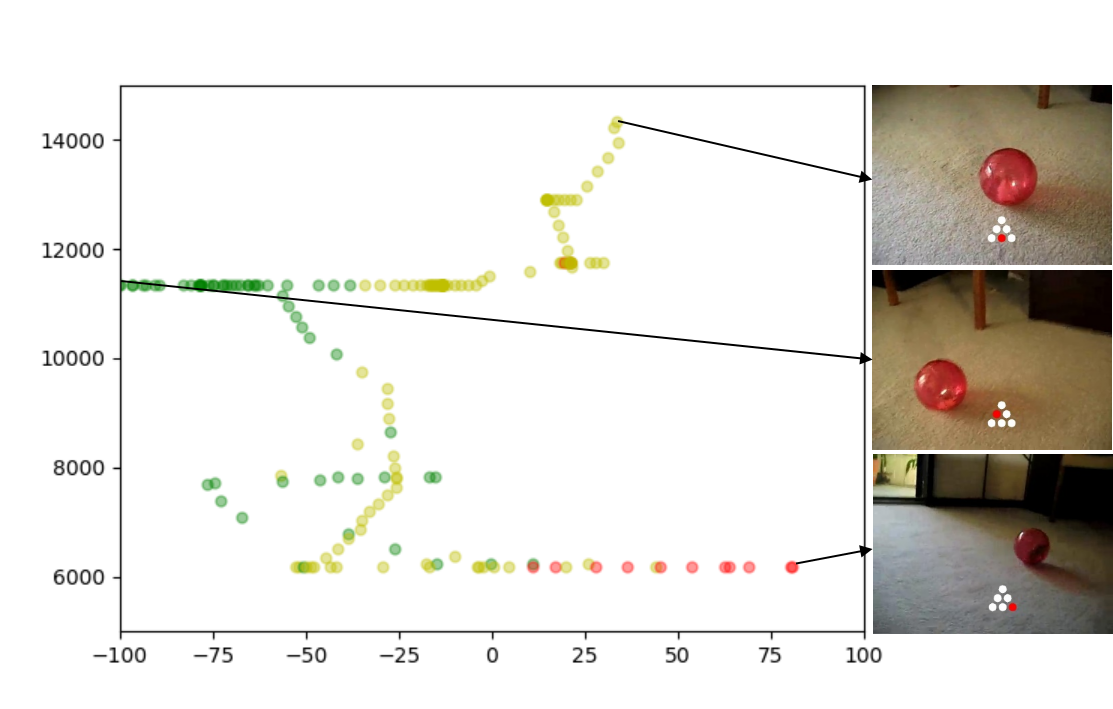} 
\vspace{-0.5cm}
\caption{Actions output from the proposed active tracker of the Woman (top) and Sphere (bottom) sequences.}
\label{fig:woman_sphere}
\vspace{-0.6cm}
\end{figure}

\subsection{Active Tracking in The UE Environment}
\label{subsec:ue_exp}
We firstly compare models trained with randomized environment and single environment. Then we test our active tracker in four environments and also compare it against traditional trackers.

\textbf{RandomizedEnv versus SingleEnv.} Based on the \emph{Square} environment, we train two models individually by the \emph{RandomizedEnv} protocol (random number of invisible background objects and starting point) and \emph{SingleEnv} protocol (fixed environment). They are tested in the \emph{S2MP2} environment, where the map, target, and the path are unseen during training. As shown in Tab. \ref{table:diff_protocol_ue}, similar results are obtained as those in Tab. \ref{table:diff_protocol}. We believe that the improvement benefits from the environment randomness brought by the proposed environment augmentation techniques. In the following, we only report the results of \emph{RandomizedEnv} protocol.

\begin{table}
\caption{Performance of different protocols in \emph{S2MP2}.}
\label{table:diff_protocol_ue}
\centering
{\begin{tabular}{c c c c}
\hline
\textbf{Protocol} & \textbf{AR} & \textbf{EL}\\
\hline
\emph{RandomizedEnv}   &1203.6$\pm$1428.4 & 2603.6$\pm$503.0 \\
\emph{SingleEnv}    & -453.4$\pm$1.5  & 461.9$\pm$180.0 \\
\hline
\end{tabular}
}
\vspace{-0.4cm}
\end{table}

\begin{figure}[t]
\begin{center}
\includegraphics[width=0.9\linewidth]{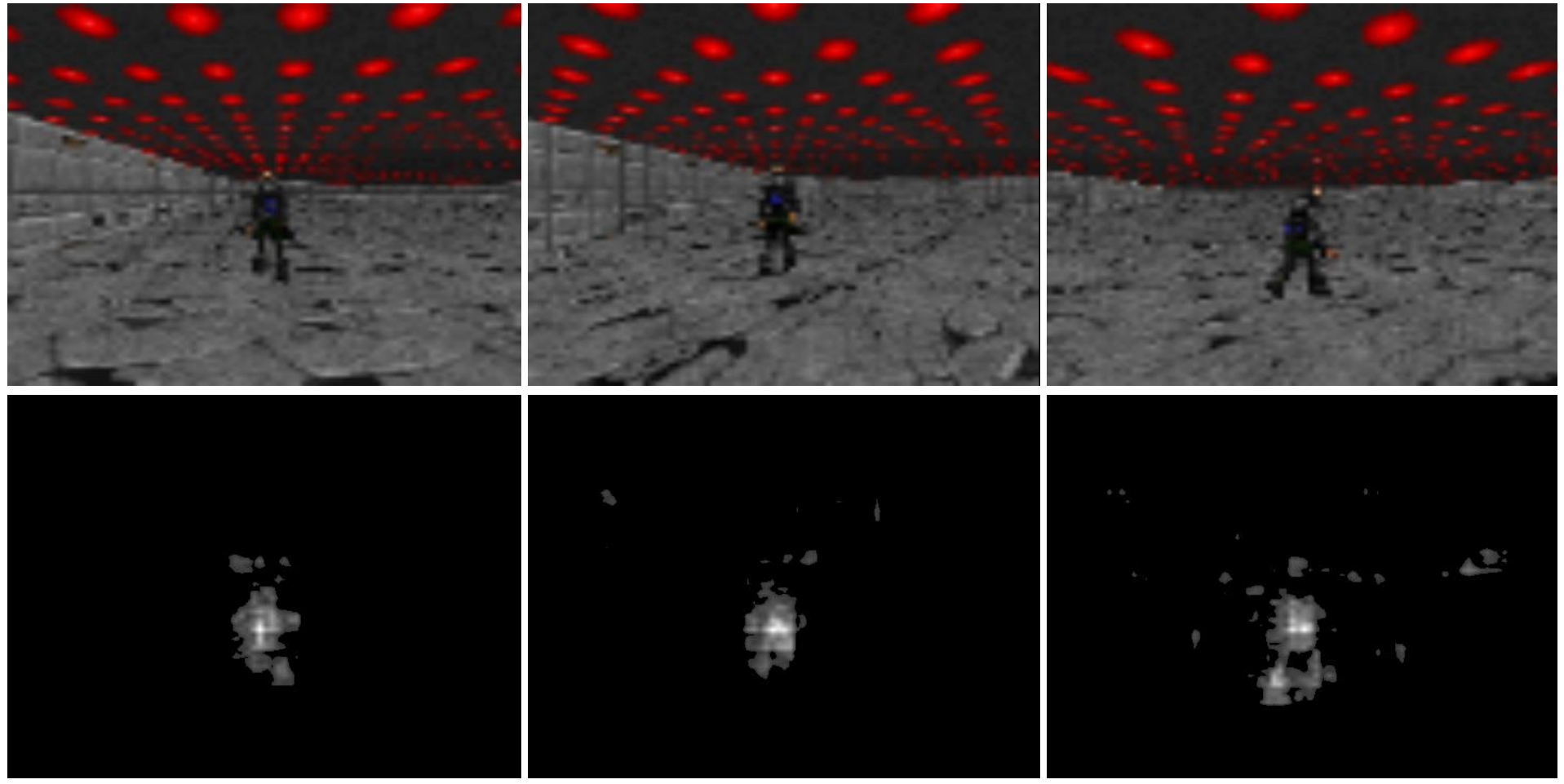}
\end{center}
\vspace{-0.3cm}
\caption{Saliency maps learned by the tracker. The top row shows input observations, and the bottom row shows their corresponding saliency maps. The corresponding actions of these input images are \emph{turn-right-and-move-forward, turn-left-and-move-forward} and \emph{turn-left-and-move-forward}, respectively. These saliency maps clearly show the focus of the tracker.}
\label{fig:saliency}
\vspace{-0.5cm}
\end{figure}

\textbf{Various Testing Environments.} 
To intensively investigate the generalization ability of the active tracker, we test it in four different environments and present the results in Tab. \ref{table:comparison_thirdparty_ue}. 
We compare it with the simulated active trackers described in Sec. \ref{subsec:vizdoom_exp},
as well as one based on the long-term TLD tracker \cite{kalal2012tracking}.

 
According to the results in Tab. \ref{table:comparison_thirdparty_ue} we conduct the following analysis:
1) Comparison between \emph{S1SP1} and \emph{S1MP1} shows that the tracker generalizes well even when the model is trained 
with target Stefani, revealing that it does not overfit to a specialized appearance.
2) The active tracker performs well when changing the path (\emph{S1SP1} versus \emph{S1SP2}), demonstrating that it does not act by memorizing specialized path.
3) When we change the map, target, and path at the same time (\emph{S2MP2}), though the tracker could not seize the target as accurately as in previous environments (the AR value drops), it can still track objects robustly (comparable EL value as in previous environments), proving its superior generalization potential.
4) In most cases, the proposed tracker outperforms the simulated active tracker, or achieves comparable results if it is not the best. The results of the simulated active tracker also suggest that it is difficult to tune a unified camera-control module for them, even when a long term tracker is adopted (see the results of TLD). However, our work exactly sidesteps this issue by training an end-to-end active tracker.

\begin{table}[t]
\caption{Comparison with traditional trackers. The best results are shown in bold.}
\label{table:comparison_thirdparty_ue}
\centering
\setlength\tabcolsep{1.5pt}
{\begin{tabular}{c  c  c  c}
\hline
\textbf{Environment} & \textbf{Tracker} & \textbf{AR} & \textbf{EL}\\
\hline
\multirow{5}{*}{\emph{S1SP1}}
& MIL        & -453.8 $\pm$0.8 &  601.4 $\pm$ 300.9  \\
& Meanshift  & -454.1$\pm$1.3 &  628.6$\pm$111.2    \\
& KCF         & -453.6$\pm$2.5 &  782.4$\pm$136.1  \\
& Correlation & -454.9$\pm$0.9 & 1710.4$\pm$417.0  \\
& TLD & -453.6$\pm$1.3 & 376.0$\pm$70.9  \\
\cline{2-4}
& Active   & \textbf{2495.7}$\pm$\textbf{12.4} &  \textbf{3000.0}$\pm$\textbf{0.0} \\
\hline
\multirow{5}{*}{\emph{S1MP1}}
& MIL        & -358.7$\pm$189.4 &  1430.0$\pm$825.3   \\
& Meanshift   & 708.3$\pm$3.3 &  \textbf{3000.0}$\pm$\textbf{0.0}   \\
& KCF         & -453.6$\pm$2.6 & 797.4$\pm$37.5    \\
& Correlation & -453.5$\pm$1.3 & 1404.4$\pm$131.1  \\
& TLD & -453.4$\pm$2.0 & 651.0$\pm$54.5  \\
\cline{2-4}
& Active   & \textbf{2106.0}$\pm$\textbf{29.3}  & \textbf{3000.0}$\pm$\textbf{0.0} \\
\hline
\multirow{5}{*}{\emph{S1SP2}}
& MIL         & -452.4$\pm$0.7 &  420.2$\pm$104.9   \\
& Meanshift   & -453.0$\pm$1.8  &    630.2$\pm$223.8 \\
& KCF         & -453.9$\pm$1.5 &  594.0$\pm$378.8   \\
& Correlation & -452.4$\pm$0.4 &  293.8$\pm$97.4  \\
& TLD & -454.7$\pm$1.8 & 218.0$\pm$26.0  \\
\cline{2-4}
& Active   & \textbf{2162.5}$\pm$\textbf{48.5}  & \textbf{3000.0}$\pm$\textbf{0.0}  \\

\hline
\multirow{5}{*}{\emph{S2MP2}}
& MIL         & -453.1$\pm$0.9 &  749.0$\pm$301.0   \\
& Meanshift   & 726.5$\pm$10.8  &  \textbf{3000.0}$\pm$\textbf{0.0}   \\
& KCF         & -452.4$\pm$1.0 &  247.8$\pm$18.8   \\
& Correlation & -215.0$\pm$475.3 & 1571.6$\pm$919.1  \\
& TLD & -453.1$\pm$1.8 & 208.8$\pm$33.1  \\
\cline{2-4}
& Active   &  \textbf{740.0}$\pm$\textbf{577.4} & 2565.3$\pm$339.3  \\
\hline
\end{tabular}
}
\vspace{-0.5cm}
\end{table}

\subsection{Active Tracking in Real-world Scenarios}
\label{subsec:real-world-exp}
To evaluate how the active tracker performs in real-world scenarios, we take the network trained in a UE environment and test it on a few video clips from the VOT dataset \cite{VOT_TPAMI}. Obviously, we can by no means control the camera action for a recorded video. 
However, we can feed in the video frame sequentially and observe the output action predicted by the network, checking whether it is consistent with the actual situation.

Fig. \ref{fig:woman_sphere} shows the output actions for two video clips named Woman and Sphere, respectively. 
The horizontal axis indicates the position of the target in the image, with a positive (negative) value meaning that a target in the right (left) part. 
The vertical axis indicates the size of the target, \ie, the area of the ground truth bounding box. Green and red dots indicate turn-left/turn-left-and-move-forward and turn-right/turn-right-and-move-forward actions, respectively. Yellow dots represent No-op action. As the figure show, 
1) When the target resides in the right (left) side, the tracker tends to turn right (left), trying to move the camera to ``pull'' the target to the center. 2) When the target size becomes bigger, which probably indicates that the tracker is too close to the target, the tracker outputs no-op actions more often, intending to stop and wait the target to move farther. 

We believe that the qualitative evaluation shows evidence that the active tracker, learned from purely the virtual environment, is able to output correct actions for camera control in real-world scenarios. 
Due to the constraint of space, we include more results of the real-world scenarios in the supplementary materials.

\subsection{Action Saliency Map}
\label{subsec:saliency}
We are curious about what the tracker has learned so that it leads to good performance. To this end, we follow the method in \cite{simonyan2013deep} to generate a saliency map of the input image with regard to a specific action. Making it more specific, an input frame $s_i$ is fed into the tracker and forwarded to output the policy function. An action $a_i$ will be sampled subsequently. Then the gradient of $a_i$ with regard to $s_i$ is propagated backwards to the input layer, and a saliency map is generated. This process calculates exactly which part of the original input image influences the corresponding action with the greatest magnitude.  

Note that the saliency map is image specific, \ie, for each input image a corresponding saliency map can be derived. Consequently, we can observe how the input images influence the tracker's actions. Fig. \ref{fig:saliency} shows  a few pairs of input image and corresponding saliency map. The saliency maps consistently show that the pixels corresponding to the object dominate the importance to actions of the tracker. It indicates that the tracker indeed learns how to find the target.

\section{Conclusion}
We proposed an end-to-end active tracker via deep reinforcement learning. Unlike conventional passive trackers, the proposed tracker is trained in simulators, saving the efforts of human labeling or trail-and-errors in real-world. It shows good generalization to unseen environments. The tracking ability can potentially transfer to real-world scenarios. 

\subsubsection*{Acknowledgement}
We appreciate the anonymous ICML reviews that improve the quality of this paper. Thank Jia Xu for his helpful discussion. Fangwei Zhong and Yizhou Wang were supported in part by the following grants 973-2015CB351800, NSFC-61625201, NSFC-61527804.
\newpage
\small\bibliography{example_paper}
\bibliographystyle{icml2018}

\end{document}


\twocolumn[
\icmltitle{Supplementary Materials for ``End-to-end Active Object Tracking via Reinforcement Learning''}



\icmlsetsymbol{equal}{*}

\begin{icmlauthorlist}
\icmlauthor{Wenhan Luo}{equal,tai}
\icmlauthor{Peng Sun}{equal,tai}
\icmlauthor{Fangwei Zhong}{pku}
\icmlauthor{Wei Liu}{tai}
\icmlauthor{Tong Zhang}{tai}
\icmlauthor{Yizhou Wang}{pku}
\end{icmlauthorlist}

\icmlaffiliation{tai}{Tencent AI Lab}
\icmlaffiliation{pku}{Peking University}

\icmlcorrespondingauthor{Wenhan Luo}{whluo.china@gmail.com}
\icmlcorrespondingauthor{Peng Sun}{pengsun000@gmail.com}
\icmlcorrespondingauthor{Fangwei Zhong}{zfw@pku.edu.cn}
\icmlcorrespondingauthor{Wei Liu}{wl2223@columbia.edu}
\icmlcorrespondingauthor{Tong Zhang}{tongzhang@tongzhang-ml.org}
\icmlcorrespondingauthor{Yizhou Wang}{yizhou.Wang@pku.edu.cn}

\icmlkeywords{Machine Learning, ICML}

\vskip 0.3in
]



\printAffiliationsAndNotice{\icmlEqualContribution} 

\section{Introduction}
In this supplementary material,
we first give the implementation details of the proposed method.
We then present the experimental results which are not included in the main texts due to space limitation. Particularly, more visual results with regard to video clips of Woman and Sphere will be provided in Sec. \ref{woman_sphere}. Sec. \ref{other_clips} provides results of the proposed tracker on other video clips from the VOT dataset. The PID-like camera control module we developed for the simulated active tracker with the traditional trackers is illustrated in Sec. \ref{camera_control}. Additional notes of ViZDoom experiments are presented in Sec. \ref{additional_notes}. 

\section{Implementation Details}
The network parameters are updated with Adam optimization, with the initial learning rate $\alpha = 0.0001$ . The regularizer factor $\beta = 0.01$ and the reward discount factor $\gamma = 0.99$. The parameter updating frequency is $T = 20$, and the maximum global iteration for training is $100\times10^6$. Validation is performed every 70 seconds and the best validation network model is applied to report performance in testing environments. 

\section{More Results of Woman and Sphere} 
\label{woman_sphere}
\begin{figure*}[t]
\hspace*{0.06\linewidth} Forward \mySkip{0.2} Left  \mySkip{0.2} Right \mySkip{0.2} Stop \\
\includegraphics[width=\picwFour]{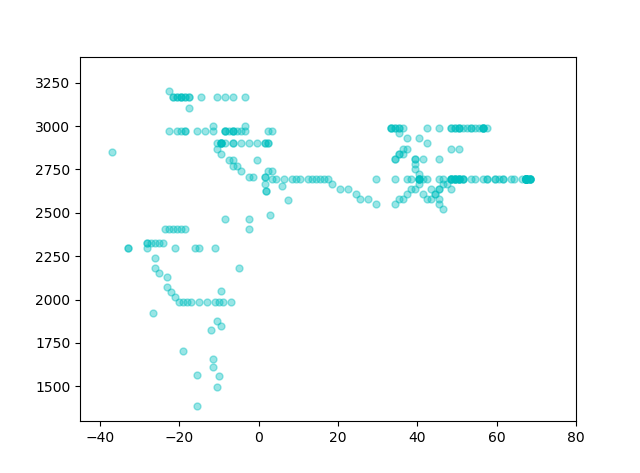}
\includegraphics[width=\picwFour]{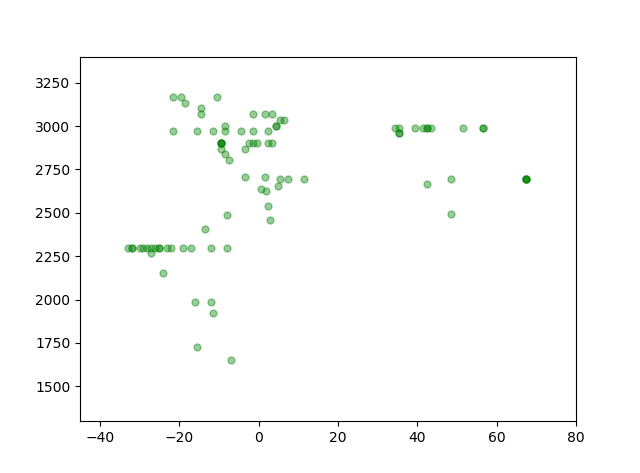}
\includegraphics[width=\picwFour]{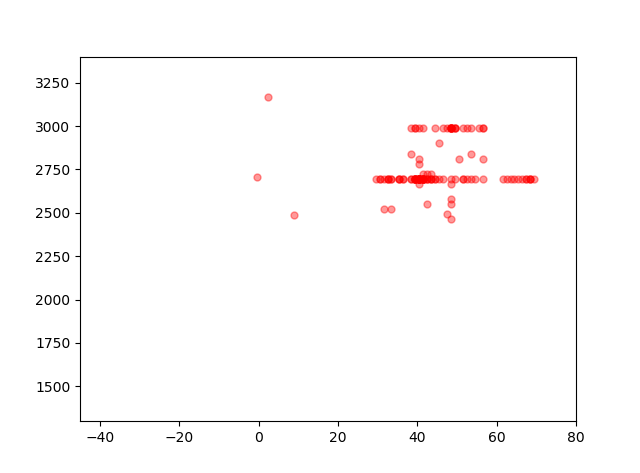}
\includegraphics[width=\picwFour]{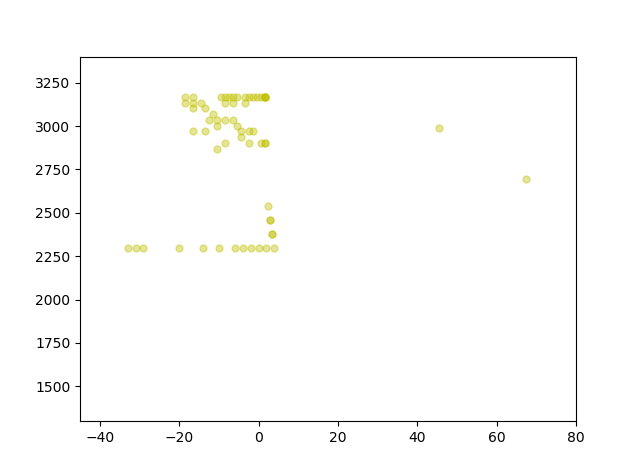} 
\caption{Visual results of individual actions of the proposed active tracker on the video clip of Woman. From left to right, they are actions of Forward (move-forward), Left (turn-left/turn-left-and-move-forward), Right (turn-right/turn-right-and-move-forward) and Stop (no-op).}
\label{fig:woman}
\end{figure*}

\begin{figure*}[th]
\hspace*{0.06\linewidth} Forward \mySkip{0.2} Left  \mySkip{0.2} Right \mySkip{0.2} Stop \\
\includegraphics[width=\picwFour]{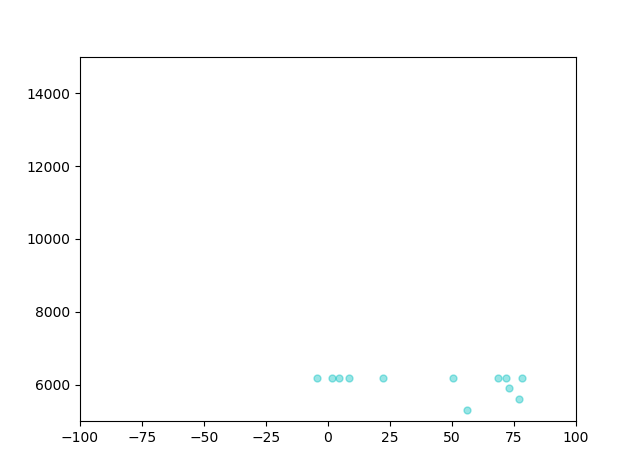}
\includegraphics[width=\picwFour]{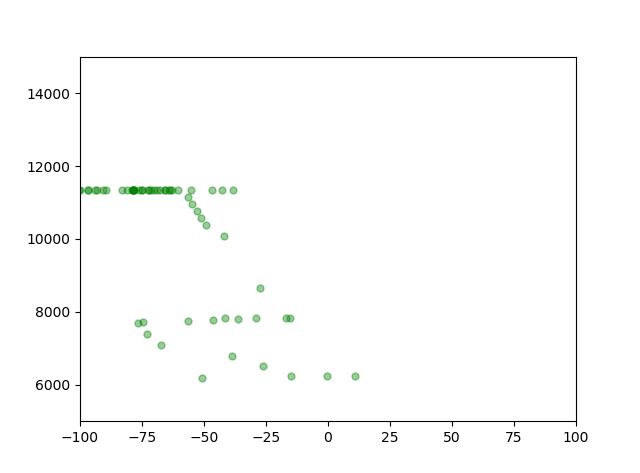}
\includegraphics[width=\picwFour]{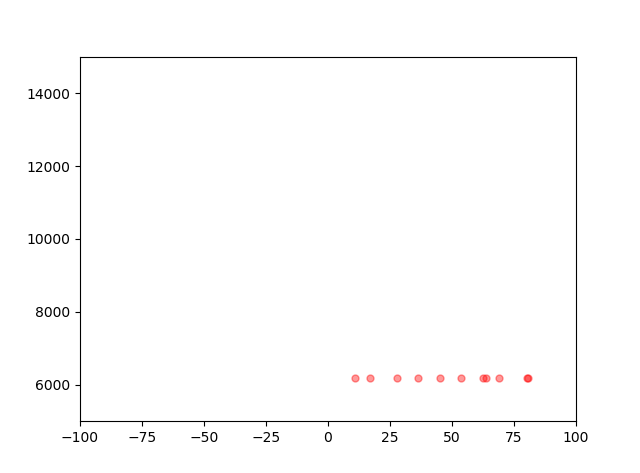}
\includegraphics[width=\picwFour]{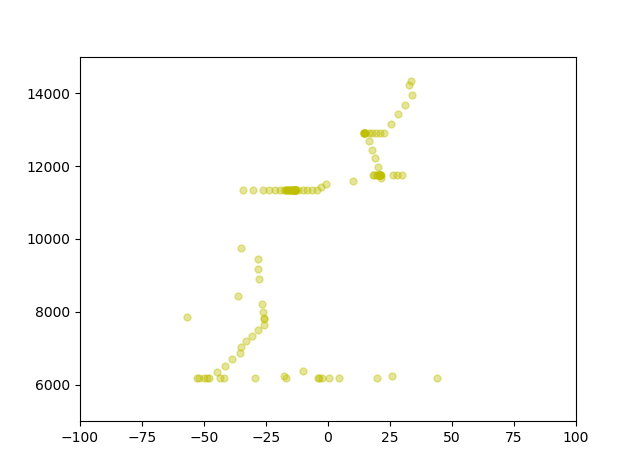} 
\caption{Visual results of individual actions of the proposed active tracker on the video clip of Sphere. From left to right, they are actions of Forward (move-forward), Left (turn-left/turn-left-and-move-forward), Right (turn-right/turn-right-and-move-forward) and Stop (no-op).}
\label{fig:sphere}
\end{figure*}

Figures \ref{fig:woman} and \ref{fig:sphere} show actions individually according to our discrete actions. The actions are grouped as Forward (Move-forward), Left (including both Turn-left and Turn-left-and-move-forward actions in our action space), Right (including both Turn-right and Turn-right-and-move-forward actions in our action space), and Stop (No-op). By doing so, the results can better indicate the potential of our tracker in real-world scenarios. Though trained in pure virtual environments, it can predict the correct actions to control the camera, further ``pulling'' the target to the center of the image.

\begin{figure}[th]
\begin{center}
\includegraphics[width=0.85\linewidth]{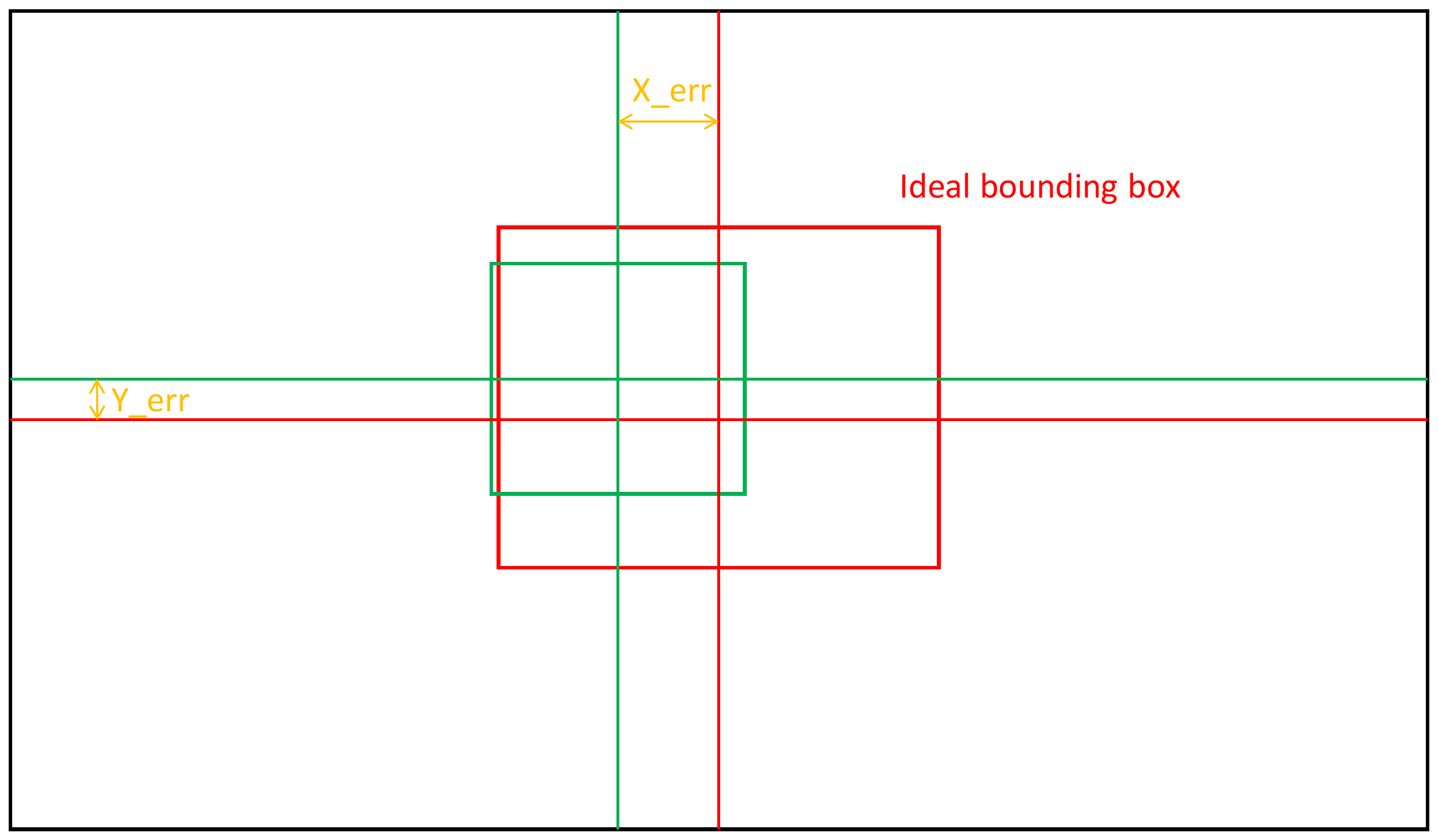}
\end{center}
\caption{An example to illustrate errors.}
\label{fig:pid}
\end{figure}

\begin{figure*}[h]
\centering
\hspace*{0.01\linewidth} Book \mySkip{0.4} Action Map  \\
\includegraphics[width=\picwTwo]{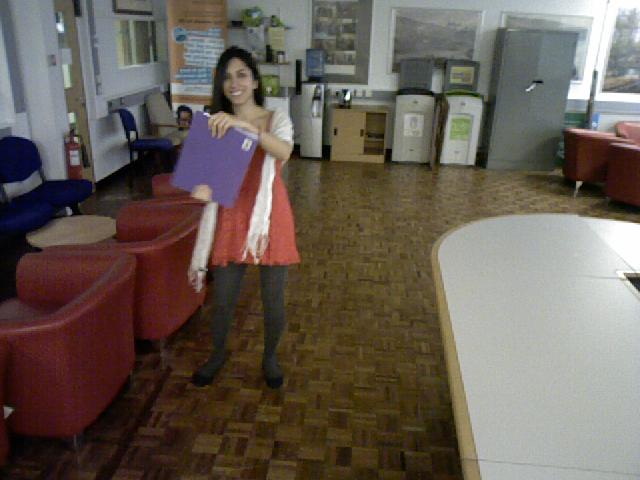}
\includegraphics[width=\picwTwo]{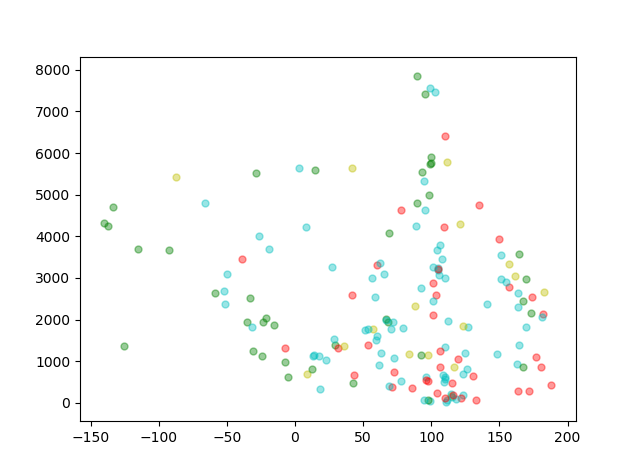}
\caption{Left: an exemplar image of video ``Book''. Right: actions output by our tracker. }
\label{fig:book}
\end{figure*}

\begin{figure*}[h]
\centering
\hspace*{0.01\linewidth} Girl \mySkip{0.4} Action Map  \\
\includegraphics[width=\picwTwo]{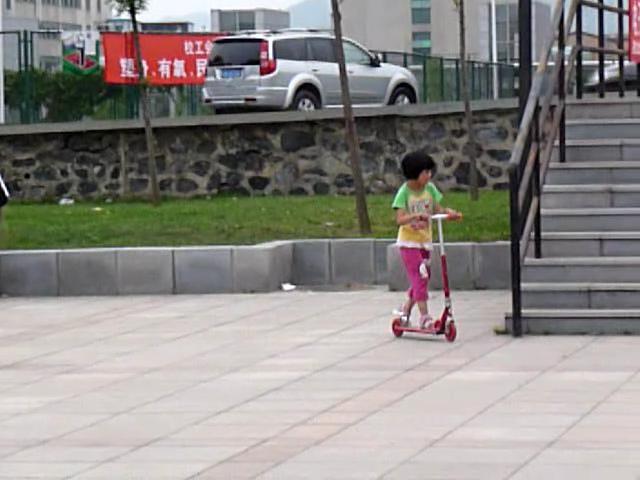}
\includegraphics[width=\picwTwo]{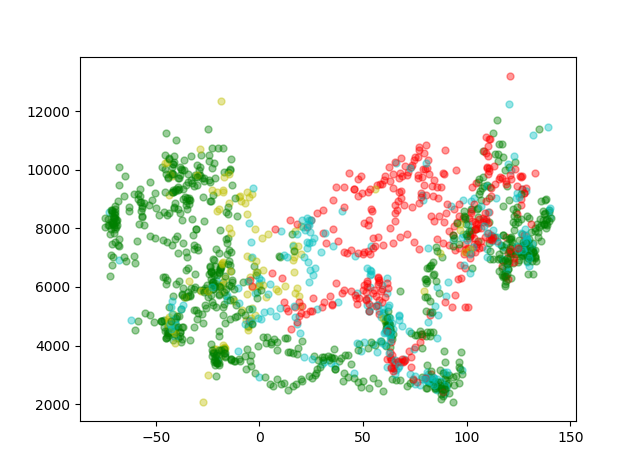}
\caption{Left: an exemplar image of video ``Girl''. Right: actions output by our tracker. }
\label{fig:girl}
\end{figure*}

\section{Results of Other Video Clips}
\label{other_clips}
\begin{figure*}[th]
\centering
\hspace*{0.01\linewidth} Iceskater \mySkip{0.4} Action Map  \\
\includegraphics[width=\picwTwo]{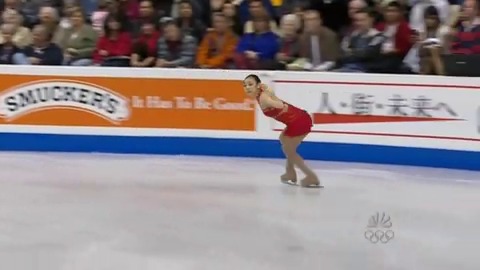}
\includegraphics[width=\picwTwo]{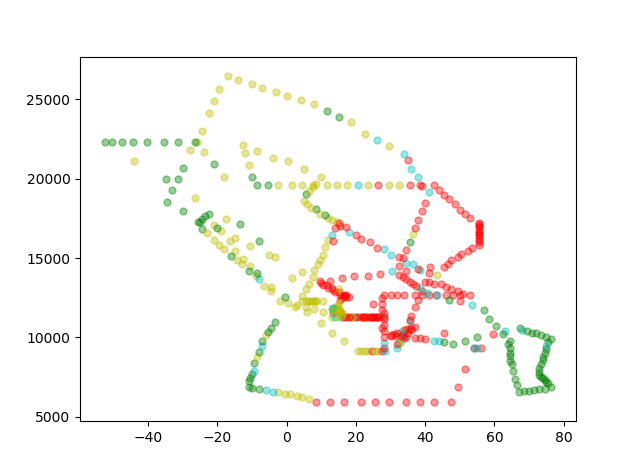}
\caption{Left: an exemplar image of video ``Iceskater''. Right: actions output by our tracker. }
\label{fig:iceskater}
\end{figure*}

\begin{figure*}[th]
\centering
\hspace*{0.01\linewidth} Iceskater1 \mySkip{0.4} Action Map  \\
\includegraphics[width=\picwTwo]{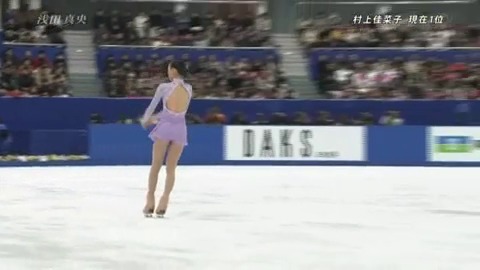}
\includegraphics[width=\picwTwo]{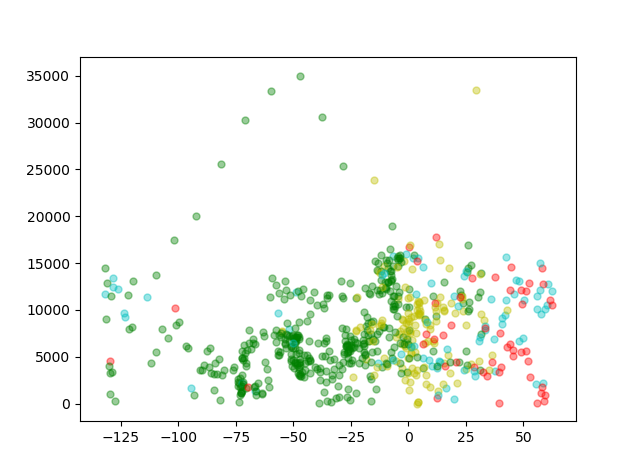}
\caption{Left: an exemplar image of video ``Iceskater1''. Right: actions output by our tracker. }
\label{fig:iceskater1}
\end{figure*}

Showing its potential even trained in virtual UE4 environment, we again test it intensively on other video clips from the VOT dataset. Figures from \ref{fig:book} to \ref{fig:iceskater1} show the actions output by the active tracker on videos named \textit{Book}, \textit{Girl}, \textit{Iceskater} and \textit{Iceskater1}.

As the same as the main submission file, green dots indicate actions of Turn-left or Turn-left-and-move-forward, and red dots represent actions of Turn-right or Turn-right-and-move-forward. Yellow dots indicate the No-op action. Blue dots respond to the Move-forward action. All the left and right actions are predicted correctly regarding the position of the target in the image (horizontal axis) except that the tracker predicts to turn left while the target is in the right part of the image (horizontally ranging from 50 to 150 in Fig. \ref{fig:girl} and horizontally ranging from 60 to 80 in Fig. \ref{fig:iceskater}). In the case of \textit{Iceskater} the incorrect predictions are made when the size of the target is very small. At that moment the camera is actually shooting a long-range perspective. Thus the target is not so salient considering the audiences, which may lead to the incorrect predictions. The gap between real-world scenarios and photo-realistic virtual environments can also result in such failures.



\section{PID-like Camera Control}
\label{camera_control}
The PID camera controller we developed is similar to a Proportional-integral-derivative controller. It decides an action sequence based on the specific error.

As shown in Fig. \ref{fig:pid}, we assume that the optimal tracking means that the rectangle is in the center of the image and takes about 20\% of the pixel space (the red bounding box in the figure). This means that we should control the camera to meet these conditions.

Formally, we denote the image width and height by $W$ and $H$, and the ideal position by $(X_{ideal}, Y_{ideal})$, respectively. We have the following parameters,

\begin{equation}
\label{eq1}
\begin{split}
X_{ideal} = W/2, \\
Y_{ideal} = H/2, \\
A_{ideal} = W\ast H\ast 20\%,
\end{split}
\end{equation}
where $A_{ideal}$ is the ideal area the object bounding box takes.

Let us assume that the result of a tracker is characterized by four parameters $\{X_{bounding}, Y_{bounding},$
$W_{bounding}, H_{bounding}\}$. Then we have the following errors,
\begin{equation}
\label{eq2}
\begin{split}
X_{err} = X_{bounding} - X_{ideal}, \\
Y_{err} = Y_{bounding} - Y_{ideal}, \\
Z_{err} = A_{ideal} - (W_{bounding}\ast H_{bounding}).
\end{split}
\end{equation}
Note that, $Z_{err}$ (depth) is tied to the size of the object. Intuitively, $X_{err}$ measures how far the bounding box is horizontally away from the desired position. In the case of Fig. \ref{fig:pid}, $X_{err}$ is negative, meaning that the camera should move left so the object is closer to the center of the image.

In general we have the following commands,

$X_{err} < 0$ means that the tracker is to the right and needs to turn left (decreasing $X$).

$X_{err} > 0$ means that the tracker is to the left and needs to turn right (increasing $X$).

$Y_{err} < 0$ means that the tracker is too far away and needs to speed up (decreasing $Y$).

$Y_{err} > 0$ means that the tracker is too close and needs to slow down (increasing $Y$).

$Z_{err} < 0$ means that the tracker is too close and needs to slow down.

$Z_{err} > 0$ means that the tracker is too far away and needs to speed up.

Note that, the action space we adopted does not include actions like ``look up'' or ``look down''. At the same time, in the view of an ideal tracker, the moving forward of the target will lead to the decrease of $Y$ value, so we intuitively map the change of $Y_{err}$ to the status of distance between the target and the tracker.

Observing this, we then map these intuitive commands to discrete actions and send actions to the environment (ViZDoom or UnrealCV). The quantity associated with a specific action depends on the quantity of a specific error. For example, if $X_{err} > 0$ and it is too large, then the quantity of turning right is also large.    

This creates a procedure which sounds a lot like a PID controller: observe a state, figure out an error and create a control sequence to reduce the error, and then repeat the process over and over again.

\section{Additional Notes}
\label{additional_notes}

\subsection{Description of Various Testing Environments of ViZDoom}
\label{descri_vizdoom}
Detailed description of the various ViZDoom testing environments are as follows. Specifically, they are obtained by modifying the \emph{Standard} environment in the following aspects:

1) Change the appearance of the target. Specifically, we have two environments named \emph{CacoDemon} and \emph{Zombie} with targets \emph{CacoDemon} and \emph{Zombie}, respectively. 

2) Revise the background in the environment. We have an environment named \emph{FloorCeiling} with different textures in ceiling and floor, and an environment named \emph{Corridor} with a corridor structure. 

3) Modify the path. The path in \emph{SharpTurn} is composed of several sharp acute angles while clockwise path is changed to a counterclockwise one in \emph{Counterclockwise}. 

4) Add distractions. \emph{Noise1} is formed by placing a same monster (stationary) near the path along which the target walks. \emph{Noise2} is almost the same as \emph{Noise1}, except that the distracting monster is closer to the path.